\documentclass[11pt]{article}

\usepackage{arxiv}
\renewcommand{\undertitle}{}
\renewcommand{\headeright}{}
\makeatletter
\renewcommand{\section}{\@startsection{section}{1}{\z@}%
  {-1.0ex \@plus -0.3ex \@minus -0.2ex}{0.3ex \@plus 0.2ex}{\large\bf\raggedright}}
\renewcommand{\subsection}{\@startsection{subsection}{2}{\z@}%
  {-0.8ex \@plus -0.3ex \@minus -0.2ex}{0.1ex}{\normalsize\bf\raggedright}}
\makeatother
\usepackage{natbib}
\usepackage{times}
\usepackage{booktabs}
\usepackage{array}
\usepackage{graphicx}
\usepackage{placeins}
\usepackage{amsmath,amsfonts,bm}

\def\eqref#1{equation~\ref{#1}}
\def\1{\bm{1}}

\def\vmu{{\bm{\mu}}}
\def\vtheta{{\bm{\theta}}}

\def\vc{{\bm{c}}}

\def\vx{{\bm{x}}}

\def\vz{{\bm{z}}}

\def\mX{{\bm{X}}}

\def\mZ{{\bm{Z}}}

\DeclareMathAlphabet{\mathsfit}{\encodingdefault}{\sfdefault}{m}{sl}
\SetMathAlphabet{\mathsfit}{bold}{\encodingdefault}{\sfdefault}{bx}{n}

\newcommand{\E}{\mathbb{E}}

\newcommand{\R}{\mathbb{R}}

\usepackage{hyperref}
\usepackage{url}
\usepackage{comment}
\title{Continual Learning without Continual Training}

\def\vsigma{{\bm{\sigma}}}
\def\vphi{{\bm{\phi}}}
\def\vpsi{{\bm{\psi}}}

\def\eqref#1{(\ref{#1})}

\author{Nikita Narayanan, Ritham Majumdarr, Sonali Parbhoo\\ Imperial College London
}

\begin{document}

\maketitle

\begin{abstract}
Continual learning requires models to adapt to new domains and new classes while retaining prior knowledge. Many existing methods rely on continued optimization, using regularization, replay, or parameter expansion to prevent new updates from overwriting previously learned knowledge. Instead, we propose replacing continual training with \emph{continual inference}: a PFN-based model that is meta-trained, and then frozen, adapting to new classes only by extending an in-context evidence set. Our model, Latent Concept PFN, performs in-context Bayesian inference over a latent concept space that captures semantic structure shared across domains and classes. As each new domain or class arrives, exemplars are added to the memory; adaptation reflects updated posterior beliefs over latent concepts rather than gradient updates. No parameters are changed, reducing forgetting. The same method handles both domain and class incremental continual learning without task identity. Concept annotations are only used during meta-training, acting as a soft anchor on the latent space rather than a fixed bottleneck. Unlike fixed-vocabulary concept methods, the model also handles noisy, ambiguous, or incomplete annotations by combining concept labels with raw input evidence to discover distinctions beyond the predefined concept set. Experiments on class and domain incremental learning datasets demonstrate competitive continual learning performance while learning interpretable latent concepts.
\end{abstract}

\section{Introduction}
\label{sec:Introduction}

Machine learning systems deployed in the real world rarely operate under stationary conditions. Clinical AI models trained at one hospital can fail when deployed elsewhere due to differences in scanners, imaging protocols, and patient populations \citep{finlayson2021clinician}. In this case the input distribution changes while the underlying prediction task remains the same, giving rise to \emph{domain-incremental} continual learning. In other cases the set of labels itself grows: a diagnostic model may need to recognise a newly defined disease or subtype while still distinguishing the conditions it was originally trained on, giving rise to \emph{class-incremental} continual learning. The two often occur together. During the COVID-19 pandemic, chest X-ray models had to accommodate a new diagnostic class while the scanners, protocols, and patient populations generating the images varied across hospitals and over the course of the pandemic \citep{degrave2021ai, roberts2021common}. In both settings, models must sequentially adapt to new data while retaining performance on what was seen before, and at test time it is typically unknown which step (and therefore which domain or classes were available) a query comes from. The central challenge is therefore twofold: models must remain plastic enough to learn from new data, while remaining stable enough to avoid catastrophic forgetting, the tendency of gradient-based training to overwrite previously learned representations when parameters are updated on new data \citep{mccloskey1989catastrophic, goodfellow2015empirical}.

Existing continual learning methods approach this challenge through parameter updates. Regularisation methods e.g. Elastic Weight Consolidation \citep{kirkpatrick2017overcoming}, Synaptic Intelligence \citep{pmlr-v70-zenke17a} penalise changes to parameters deemed important for previous tasks, but largely fail when no task identity is given at test time \citep{vandeven2022three}. Replay methods \citep{NEURIPS2019_fa7cdfad, NEURIPS2020_b704ea2c, chaudhry2019tiny, Rebuffi_2017_CVPR} interleave historical data with new examples during training. Class-incremental methods additionally correct the classifier's bias towards recently seen classes \citep{8954008, Zhao_2020_CVPR} or allocate additional parameters as new classes arrive \citep{rusu2022progressive, Mallya_2018_ECCV, Yan_2021_CVPR}. Methods built on large pretrained models instead learn a small set of prompts on top of frozen features \citep{Wang_2022_CVPR, wang2022dualprompt} or train a new set of low-ranked parameters when a new task arrives \citep{liang2024inflora, wu2024sdlora}.

More recently, concept-based methods \citep{Yu_2025_CVPR, javadi2026cicbm} have been used to provide interpretable structure in continual learning, however these methods still depend on explicit concept supervision and sequential weight updates across new tasks. Although different in mechanism, these approaches all share a common premise: \emph{adaptation requires further optimisation.} As a result, they inherit the fundamental tension between plasticity i.e. learning from new data, and stability i.e. retaining prior knowledge, with catastrophic forgetting arising whenever new updates interfere with previously learned representations. Prototype-based methods are a partial exception: they classify by comparing inputs to class prototypes in the feature space of a frozen pretrained model \citep{NEURIPS2023_2793dc35, zhou2025revisiting}, and so largely avoid updating weights on the stream. However, each class is summarised by a single set of statistics in a fixed feature space, which cannot represent a class whose relationship to the inputs differs across domains.

In this paper, we argue that this framing is unnecessarily restrictive. When domains evolve or new classes arrive, the quantity that must be preserved is \emph{not a specific parameter configuration}, but the conceptual structure that governs the mapping from observations to labels. A rotated digit retains its identity, and a pathology remains characterised by the same morphology despite variation in imaging conditions. Many of these concepts are also shared across classes: the vertical and horizontal strokes that make up handwritten letters also make up digits. From this perspective, continual learning is fundamentally not a problem of preserving model weights, but rather of inferring stable latent concepts from changing observations and relating them to the labels seen so far. This perspective suggests replacing continual training with \emph{continual inference}. If a model can infer the shared latent concepts underlying each new domain or class, then adaptation can occur through belief updating rather than gradient descent. This would avoid parameter overwriting entirely, while allowing knowledge acquired earlier in the stream to transfer naturally whenever latent concepts are reused. 

Prior-data fitted networks (PFNs) \citet{muller2022transformers, hollmann2023tabpfn, hollmann2025accurate} offer a natural starting point. They are trained to approximate the posterior predictive over many datasets drawn from a prior. Although PFNs were designed for prediction on a single, static dataset, we extend this to the continual learning setting and map inputs to a structured latent space anchored to human-interpretable concepts before predicting targets. We meta-train a PFN-based model once, before the continual learning stream, and then keep it frozen. New domains and classes are then learned solely by conditioning on a growing context of stored examples, so no parameters are updated and none can be overwritten.

\textbf{Contributions.} \textbf{(1)}~We introduce Latent Concept PFN, an in-context learner that maps inputs to a structured latent space shaped by concepts and then to targets. \textbf{(2)}~We show that the Latent Concept PFN achieves competitive continual learning performance with minimal forgetting, without parameter updates on the continual learning stream. \textbf{(3)}~We show that the structured latent space can be used to explain errors in the model's output. \textbf{(4)}~We show that the latent space captures information beyond the given concepts and therefore is not restricted to a fixed concept vocabulary.

\section{Related Work}

\label{sec:related_work}
\textbf{Domain-Invariant Representation Learning.}
Methods such as Domain-Adversarial Neural Networks \citep{JMLR:v17:15-239}, Invariant Risk Minimization \citep{arjovsky2020invariant}, and the Invariant Information Bottleneck \citep{li2022invariant} learn representations that are stable across domains, sharing our assumption that domain-invariant latent features exist and are sufficient for prediction. These methods require domain labels and joint optimization across domains to align representations explicitly. In our model, we instead build invariance at meta-training time by anchoring the latent space to concept annotations shared across domains and classes, and by including shift episodes in which inputs are warped while concepts and labels are held fixed, training the encoder to recover the same latent concepts despite changes in appearance. During the continual learning stream no parameters are updated, so invariance is a fixed property of the model rather than something that must be learned at deployment.

\textbf{Temporal Domain Generalization.}
Temporal domain generalization (TDG) methods characterize distributional drift to extrapolate to unseen future domains \citep{NEURIPS2023_459a911e}, whereas continual learning requires retaining performance on all previously seen domains. Existing TDG methods either access multiple source domains simultaneously or fold past domains into gradient updates: DRAIN \citep{bai2023temporal}, MISTS \citep{xie2024enhancing}, LSSAE \citep{pmlr-v162-qin22a}, and EvoS \citep{NEURIPS2023_459a911e} all do so. Closest to our setting, Drift-Resilient TabPFN \citep{NEURIPS2024_b2e2774c} also performs no gradient updates at test time, but it is a TDG method evaluated on a single in/out-of-distribution split rather than a sequential stream. It also assumes a fixed label set.

\textbf{Concept Bottleneck Models.}
CBMs \citep{pmlr-v119-koh20a} predict targets through an intermediate layer of human-interpretable concepts; extensions address incompleteness \citep{NEURIPS2022_944ecf65, Shang_2024_CVPR} and annotation cost \citep{yuksekgonul2023posthoc, oikarinen2023labelfree}. Recent work applies CBMs to continual learning \citep{Yu_2025_CVPR, javadi2026cicbm, lai2025learning}, but all still require concept supervision and parameter updates at each step. Our model requires concept supervision only during meta-training and performs no updates on the stream; predictions are made via in-context inference through a latent space that captures the annotated concepts while remaining free to encode discriminative structure beyond them.

\textbf{Streaming Continual Learning.}
\citet{lourencobridging} and CURE \citep{lee2026bounded} frame continual learning as a context management problem for frozen in-context models, reducing the stability-plasticity trade-off to which examples to keep in memory, a framing our method shares. The key difference is that both operate directly from raw inputs to outputs, whereas our memory lives in a latent concept space shared across domains and classes.

\section{Continual Learning Without Continual Training}
\label{sec:method}

Our goal is a continual learning system that can adapt to new domains and new classes without retraining or replaying past data. The central idea is to use \emph{shared concepts} as a bridge between tasks: the raw appearance of a disease may differ across hospitals, but findings such as pleural thickening remain relevant across both domains and are shared across classes. The Latent Concept PFN learns a latent space anchored to such concepts during meta-training, is then frozen, and adapts during the continual learning stream through in-context inference alone.

\subsection{Setup and Model}
\label{sec:setup}
Let $\vx$ denote an input and $y \in \{1, \dots, K\}$ its class label. We introduce a latent vector $\vz \in \R^d$ and an observed concept vector $\vc \in \{0,1\}^C$, a noisy, partial annotation of the semantic concepts relevant to $y$ (e.g.\ \emph{has a curved stroke} for digit images, or \emph{cardiomegaly} for chest radiographs). Predictions for a query $\vx^{(q)}$  are made given a labeled context set of $n$ examples, $\mathcal{D}_{\text{ctx}} = \{(\vx^{(i)}, y^{(i)})\}_{i=1}^{n}$, whose inputs $\vx^{(1)}, \dots, \vx^{(n)}$ whose inputs form the rows of $\mX_{\text{ctx}}$ and whose label set we write $\mathcal{Y}_{\text{ctx}} \subseteq \{1, \dots, K\}$. During meta-training the context is part of an episode drawn from the prior (Section~\ref{sec:metatrain}); during continual learning it is the exemplar memory (Section~\ref{sec:testtime}).

The model makes its prediction in two stages: i) A latent encoder $q_{\vtheta}$ maps each input to a posterior over a latent $z$ by attending to the other inputs in the context. ii) An in-context classifier $g_{\vpsi}$, a pretrained TabPFN \citep{hollmann2025accurate} that remains frozen, predicts the query label from the latents and the labels in the context. Concept annotations enter only through a separate concept head $h_\vphi$ (Section ~\ref{sec:concepts}) which shapes $z$ during meta-training and is not used for prediction.

\subsection{Concept and Label Inference}
\label{sec:stage1}

\textbf{Concept Inference.} The latent encoder $q_\theta$ consists of an input embedding network (e.g. a CNN for image inputs) followed by a transformer with the PFN attention mask \citep{muller2022transformers} which outputs a Gaussian posterior over the latent $z$ at every position:
\begin{equation}
  q_{\vtheta}(\vz \mid \vx, \mX_{\text{ctx}})
  = \mathcal{N}\!\big(\vz;\ \vmu_{\vtheta}(\vx, \mX_{\text{ctx}}),\ \text{diag}\big(\vsigma^2_{\vtheta}(\vx, \mX_{\text{ctx}})\big)\big).
  \label{eq:encoder}
\end{equation}
Under the mask, context positions attend to one another and each query attends to the context only, so queries do not influence each other. The parameters $\vtheta$ are the only parameters of the predictive path that are trained.

\textbf{Label Inference.}
\label{sec:stage2}
The in-context classifier is TabPFN \citep{hollmann2025accurate}, with parameters $\vpsi$ that are held fixed throughout. It receives a table with one row per context
example, containing the latent $\vz^{(i)} \sim q_{\vtheta}(\cdot \mid \vx^{(i)}, \mX_{\text{ctx}})$ drawn from Eq ~\ref{eq:encoder}
and its label $y^{(i)}$, together with the query latent $\vz^{(q)} \sim q_{\vtheta}(\cdot \mid \vx^{(q)}, \mX_{\text{ctx}})$,
and returns a distribution over the query label. TabPFN requires labels to be
consecutive integers, and assigns one probability to each. We therefore relabel the context with the rank
map $\rho_{\text{ctx}} : \mathcal{Y}_{\text{ctx}} \to \{1, \dots, |\mathcal{Y}_{\text{ctx}}|\}$, which sorts the classes
present in the context and replaces each by its position; a context containing classes
$\{3, 7, 9\}$ is relabeled $3 \mapsto 1$, $7 \mapsto 2$, $9 \mapsto 3$. Writing $\tilde g_{\vpsi}$ for TabPFN's output distribution over ranks, the distribution over the original classes is
obtained by mapping ranks back:
\begin{equation}
  g_{\vpsi}\big(y \mid \vz^{(q)}, \{(\vz^{(i)}, y^{(i)})\}_{i=1}^{n}\big) =
  \begin{cases}
    \tilde g_{\vpsi}\big(\rho_{\text{ctx}}(y) \mid \vz^{(q)}, \{(\vz^{(i)}, \rho_{\text{ctx}}(y^{(i)}))\}_{i=1}^{n}\big)
      & y \in \mathcal{Y}_{\text{ctx}}, \\
    0 & \text{otherwise}.
  \end{cases}
  \label{eq:head}
\end{equation}
Classes absent from the context receive no probability mass. During meta-training this order is randomly permuted per episode,
Section~\ref{sec:metatrain}, so that the encoder cannot exploit a fixed class ordering.

\textbf{Posterior Predictive.}
\label{sec:predictive}
Integrating over the latent $\vz^{(q)}$ of the query $q$ and of every context example $\mZ_{\text{ctx}}$ gives
\begin{equation}
\begin{split}
  q\big(y^{(q)} \mid \vx^{(q)}, \mathcal{D}_{\text{ctx}}\big)
  = \int\, & g_{\vpsi}\big(y^{(q)} \mid \vz^{(q)}, \{(\vz^{(i)}, y^{(i)})\}_{i=1}^{n}\big) \\
  & \times \underbrace{q_{\vtheta}\big(\vz^{(q)} \mid \vx^{(q)}, \mX_{\text{ctx}}\big)}_{\text{query latent}}
    \underbrace{\prod_{i=1}^{n} q_{\vtheta}\big(\vz^{(i)} \mid \vx^{(i)}, \mX_{\text{ctx}}\big)}_{\text{context latents}}\,
    d\vz^{(q)}\, d\mZ_{\text{ctx}},
\end{split}
\label{eq:predictive}
\end{equation}
where $\mZ_{\text{ctx}}$ collects the context latents $\vz^{(i)}$. We estimate Eq.~\eqref{eq:predictive} with
$S$ joint reparameterised draws of all latents, $\big(\vz^{(q,s)}, \mZ^{(s)}_{\text{ctx}}\big)$ for
$s = 1, \dots, S$, averaged in probability space:
\begin{equation}
  \hat q_S\big(y^{(q)} \mid \vx^{(q)}, \mathcal{D}_{\text{ctx}}\big)
  = \frac{1}{S} \sum_{s=1}^{S} g_{\vpsi}\big(y^{(q)} \mid \vz^{(q,s)}, \{(\vz^{(i,s)}, y^{(i)})\}_{i=1}^{n}\big).
  \label{eq:mc}
\end{equation}
The same estimator is used during meta-training and at test time, with only the context changing.

\textbf{Conditional Independence.}
\label{sec:condind} The classifier receives latents and context labels but never the inputs or the concept annotations, so by construction
\begin{equation}
  y^{(q)} \perp \big(\vx^{(q)}, \mX_{\text{ctx}}\big) \mid \big(\vz^{(q)}, \{(\vz^{(i)}, y^{(i)})\}_{i=1}^{n}\big)
  \label{eq:condind}
\end{equation} 
This is a structural property of the model rather than an assumption about the data. This means that any label-relevant information in the inputs must pass through $z$ and the likelihood term of Eq.~\eqref{eq:loss} pushes whatever the labels require into $z$, including information not captured by the concepts. (Sec.~\ref{sec:concepts}).

\subsection{Concept Anchoring during Meta-training}
\label{sec:concepts}

Importantly, the observed concept $\vc$ enters only through an auxiliary head, and does not play a role in prediction. We
attach a separate concept network $h_{\vphi} : \R^d \to [0,1]^C$, a linear map followed by an element-wise sigmoid,
to the latents and supervise it to predict $\vc^{(i)}$ from $\vz^{(i)}$. We choose the concept vocabulary to be meaningful on both the meta-training data and on the data of the continual learning stream (e.g. vertical stroke or horizontal stroke applies to both letters and digits). The auxiliary loss pulls $\vz$ towards
concept-aligned representations. As intermediate concepts tend to be stable across domains and shared across classes, the concept loss
anchors $z$ to structure that does not move when the appearance of x shifts, which we expect to support forward
transfer when later domains or classes reuse the same concepts.

Critically, $\vz$ is not constrained to $\vc$: the auxiliary loss acts as a soft anchor rather than a hard
bottleneck. The concept loss regularises $\vz$ toward concept space through the BCE term (see Eq.~\eqref{eq:loss}), but does not confine it: the residual dimensions of $\vz$ remain free to encode discriminative structure
beyond the vocabulary. This matters when two
classes share a concept code, as several EMNIST letters do in our experiments (App. \ref{app:concepts}). A concept bottleneck model cannot separate
such classes, and the concept loss provides no signal to do so. The signal comes instead from the label likelihood,
which reaches z through the classifier and rewards any feature of the input that separates the classes. Additionally, the encoder can compare its inputs with the other context inputs through attention. The latent can therefore represent
distinctions that the concept vocabulary omits or annotates incorrectly, which we examine directly in Sec. \ref{sec:results}.

\subsection{Meta-Training for Continual Inference}
\label{sec:metatrain}
Meta-training has to prepare a single frozen model for two kinds of change it will meet on the stream: new classes whose meaning must be read from the context (class-incremental), and new domains whose inputs look different but carry the same concepts (domain incremental). We build both into the prior over episodes.

\textbf{Source data and episodes.} We meta-train on a source dataset $\mathcal{S} = \{(\vx^{(j)}, \vc^{(j)}, y^{(j)})\}_{j=1}^{N}$ with concept
annotations. The source data must share the input space and concept vocabulary of the continual learning stream, but need not
contain its classes or domains. In our MNIST experiments, the source and stream labels sets are disjoint,
$\mathcal{Y}^{\text{src}} \cap \mathcal{Y} = \emptyset$, so everything a model knows about a stream class must reach it through context. Note that due to lack of available data for the CXR datasets, we could not do the same but we instead ensure that no image used in meta-training
appears in the stream. Each
episode draws a random subset of classes and splits a sample of them into a context set
$\mathcal{D}_{\text{ctx}} = \{(\vx^{(i)}, \vc^{(i)}, y^{(i)})\}_{i=1}^{n}$ and $m$ queries
$\mathcal{D}_{\text{qry}} = \{(\vx^{(q)}, \vc^{(q)}, y^{(q)})\}_{q=n+1}^{n+m}$ whose classes all appear in the context.
We write $\mathcal{D} = \mathcal{D}_{\text{ctx}} \cup \mathcal{D}_{\text{qry}}$ for the whole episode. In an identity episode, the label is the source class. We add two further episode types.

\textbf{Rule episodes (c to y).} In these episodes, the label is a random function of the concept code and to a controlled degree, the pixel inputs. We draw a network $f_\omega$ from the BNN prior of TabPFN \citep{hollmann2023tabpfn}, and apply it row by row,
\begin{equation}
y^{(i)} \;=\; \operatorname{bin}\!\Big(f_\omega\big(\big[\,2c^{(i)}-1 \;;\; \kappa\,(x^{(i)}-\bar x)\,\big]\big)\Big),
\qquad \omega \sim p(\omega),
\label{eq:rule-prior}
\end{equation}
where $\bar x$ is the mean input of the episode and $\operatorname{bin}$ sorts the real-valued outputs into
$|\mathcal Y_{\mathrm{ctx}}|$ label groups by cutting them at $|\mathcal Y_{\mathrm{ctx}}|-1$ of the episode's own
outputs. The scale $\kappa$, drawn per episode, sets how much of the rule reads the pixels rather than the concepts. For a proportion of rule episodes, $\kappa = 0$ and the label is a function of $c$ alone. Since the same concept code maps to different labels in different episodes, and the classifier sees only z,
a rule episode is answerable only if z carries the concept code and the model reads the mapping from the context. This is what a class-incremental stream demands as new classes arrive.

\textbf{Shift episodes (x to c).} In these episodes, the labels and concept annotations are those of the source class and only the inputs change. The episode draws $K \sim \mathcal U\{1,\dots,5\}$ transformations and assigns each row to one of them. Transformation $k$ is a smooth
coordinate map $g_{\xi_k} : \mathbb R^2 \to \mathbb R^2$ drawn from the same hyperprior as the rules, giving
\begin{equation}
x'^{(i)}(p) \;=\; x^{(i)}\big(p + \tau_k\,(g_{\xi_k}(p) - \bar g_{\xi_k})\big),
\qquad (x^{(i)}, c^{(i)}, y^{(i)}) \;\mapsto\; (x'^{(i)}, c^{(i)}, y^{(i)}),
\label{eq:shift-prior}
\end{equation}
where $p$ ranges over pixel coordinates. The mean displacement $\bar g_{\xi_k}$ is removed so that the content stays in the frame and $\tau_k$ is set so that the mean displacement is a fraction $\alpha_k$ of the mean pixel radius, with
$\alpha_k = 0$ for a quarter of the transformations and otherwise log-uniform on $[0.05, 0.5]$. This results in smooth non-linear warps that move strokes rather than scramble pixels so that the transformed inputs still resemble each other and the targets \citep{farquhar2019robust}. With probability $0.3$, one transformation is withheld from the context which means that its queries must be answered from a context in which every input was transformed differently. This asks the encoder to recover the
same concepts and label from an input whose appearance has changed, which is what a domain-incremental stream
demands. 

\textbf{Prior.} The prior is therefore a mixture of three kinds of episode,
\begin{equation}
p(D) \;=\; \pi_{\mathrm{rule}}\, p_{\mathrm{rule}}(D)
       \;+\; \pi_{\mathrm{shift}}\, p_{\mathrm{shift}}(D)
       \;+\; \big(1 - \pi_{\mathrm{rule}} - \pi_{\mathrm{shift}}\big)\, p_{\mathrm{id}}(D),
\label{eq:prior-mixture}
\end{equation}
where $p_{\mathrm{id}}$ draws an episode labelled by source-class identity, as described above, and the mixture
weights are given in App.~\ref{app:method}. A rule episode carries unwarped inputs and a shift episode carries class-identity labels, so the two additions never apply to the same episode.
With $\vpsi$ fixed, only the encoder parameters $\vtheta$ and the concept network parameters $\vphi$ are fitted, by
minimising
\begin{align}
  \mathcal{L}(\vtheta, \vphi)
  = {} & \E_{\mathcal{D} \sim p(\mathcal{D})} \left[ \frac{1}{m} \sum_{q=n+1}^{n+m}
        -\log \hat q_S\big(y^{(q)} \mid \vx^{(q)}, \mathcal{D}_{\text{ctx}}\big) \right] \nonumber \\
  & + \lambda_c\, \E_{\mathcal{D} \sim p(\mathcal{D})} \left[ \frac{1}{n+m} \sum_{i=1}^{n+m}
        \E_{\vz^{(i)} \sim q_{\vtheta}(\cdot \mid \vx^{(i)}, \mX_{\text{ctx}})}
        \mathrm{BCE}\big(h_{\vphi}(\vz^{(i)}),\, \vc^{(i)}\big) \right],
  \label{eq:loss}
\end{align}
where $\hat q_S$ is the estimator~\eqref{eq:mc}.

Rule episodes only change the concept-label mapping while the shift episode changes only the inputs. The concept term is applied at every position of the episode, context and query alike, since concept annotations are
available for all source examples. After meta-training, $\vtheta$ and $\vphi$ are frozen. No rules or warps drawn during the continual learning stream.

\subsection{Continual Inference}
\label{sec:testtime}

During the continual learning stream, no parameter is updated and no concept annotations are used. Continual learning is carried out entirely by conditioning on a growing exemplar memory, which serves as the context. After step $t$, a subset $\tilde{\mathcal{D}}_t \subset \mathcal{D}_t$ of $k$ labeled examples per class is drawn
uniformly at random and stored:
\begin{equation}
  \mathcal{M}_t = \mathcal{M}_{t-1} \cup \tilde{\mathcal{D}}_t, \qquad \mathcal{M}_0 = \emptyset.
  \label{eq:memory}
\end{equation}
The memory stores only input--label pairs and is never otherwise modified. For a test query $\vx^{(q)}$ after step $t$, the model predicts with \eqref{eq:mc} using the memory as the context,
$\mathcal{D}_{\text{ctx}} = \mathcal{M}_t$ (so $n = |\mathcal{M}_t|$). Predictions are made by
$\hat y^{(q)} = \arg\max_{y} \hat q_S\big(y \mid \vx^{(q)}, \mathcal{M}_t\big)$.

In the class-incremental setting, the memory label set is
$\mathcal{Y}_1 \cup \dots \cup \mathcal{Y}_t$, and by Eq.~\eqref{eq:head} classes that have not yet arrived receive no
probability mass. In the
domain-incremental setting, the memory mixes exemplars from every domain seen so far. In both cases, the memory is a mixture over steps and hence no task or domain identity is provided. The model must infer
the relevant structure from the content of $\mathcal{M}_t$ alone. By default, the whole memory is used as the context. When the memory is larger than the contexts seen during
meta-training, we can instead draw $N_{\text{sub}}$ class-balanced random subsets of $\mathcal{M}_t$ of a fixed size,
evaluate Eq.~\eqref{eq:mc} on each, and average the resulting predictives in probability space.The separation between a model fixed after meta-training and a memory that grows
with the stream is the mechanism by which the model adapts to new domains and classes without retraining.

\section{Experimental Setup}
\label{sec:experiments}

\textbf{Datasets.} We run three experiments (summarised in Table~\ref{tab:datasets}). Every experiment has two disjoint parts: a source dataset with concept annotations, used for meta-training (Section~\ref{sec:metatrain}), and a continual learning stream, which the model adapts to only through context (Section~\ref{sec:testtime}). We run two MNIST experiments which share a single source dataset and a single meta-trained model, in which one presents a domain-incremental stream \textbf{Rotated MNIST} \citep{lecun1998gradientbased, NIPS2017_f8752278}, stream of five rotations ($0^\circ$, $15^\circ$, $30^\circ$, $45^\circ$, $60^\circ$) over a fixed ten-class label space, and the other a class-incremental stream \textbf{Split MNIST} \citep{pmlr-v70-zenke17a, vandeven2022three}, introducing disjoint digit pairs $(0,1),(2,3),(4,5),(6,7),(8,9)$ in sequence. The model is meta-trained on EMNIST letters \citep{cohen2017emnist}: 26 classes of $28{\times}28$ grayscale images with $C{=}7$ binary stroke-attribute concepts with no digit classes seen during meta-training\citep{cohen2017emnist}. In both, we store $k = 5$ exemplars per class at each step, and after each step we evaluate on every task seen so far. The third experiment is on chest radiographs, \textbf{CXR}. We meta-train on a sub-set of VinDr-CXR \citep{nguyen2022vindrcxr, PhysioNet-vindr-cxr-1.0.0, pollard_physionet_2026} with a small slice (1000 images) from TBX11K \citep{liu2020rethinking}. VinDr is labelled with both local findings and global diagnoses. We use $C = 12$ local findings as our concepts (e.g.\ cardiomegaly, pleural effusion). The stream consists of four hospitals: VinDr-CXR, TBX11K \citep{liu2020rethinking, liu2023revisiting}, Shenzhen and Montgomery \citep{jaeger2014two, candemir2014lung, jaeger2014automatic}, and we predict whether or not the scan received a label of tuberculosis. The hospitals differ in various ways such as the scanners available, patient populations and the prevalence of tuberculosis, which is 5.5\%, 11\%, 50\%, and 36\% respectively in the queries. At each step we store $k = 16$ exemplars per class, so the memory holds 32 images per hospital and 128 after the last step. Full details are in Appendix~\ref{app:exptsetup}.

\textbf{Baselines.} We compare against one replay method, two pretrained-backbone continual learning methods and a standard TabPFN. i) \emph{DER++} \citep{NEURIPS2020_b704ea2c} with the replay buffer set to the size of our exemplar memory, ii) \emph{L2P} \citep{Wang_2022_CVPR} and iii) \emph{InfLoRA} \citep{liang2024inflora}, both on a ViT-B/16 backbone, pretrained on ImageNet-21k and trained for five epochs per task on the full data of each step, and iv) \emph{TabPFN} on frozen ImageNet ResNet-18 \citep{he2016deep} features projected by PCA to 16 dimensions (size of latent $z$), fitted without labels on our meta-training images and conditioned on the same memory as our method. More details in Appendices ~\ref{app:baselines_overview}, ~\ref{app:baselines}.

\textbf{Evaluation.} We use standard continual learning metrics computed over the mean accuracy $a_{ij}$ matrices: average accuracy (AA) and backward transfer (BWT) \citep{NIPS2017_f8752278}, the forgetting measure (FM) \citep{chaudhry2018riemannian} with the final step excluded from the maximum, and forward transfer (FWT) \citep{NIPS2017_f8752278}. To evaluate the concept read-out, we marginalise the concept network $h_\phi$ over the same posterior draws of $z$ that give the class prediction, and score each concept by its AUROC against annotations if available (Appendix \ref{app:evaluation_metrics}).

\section{Results}
\label{sec:results}

\begin{figure}[h]
\centering
\includegraphics[width=\linewidth]{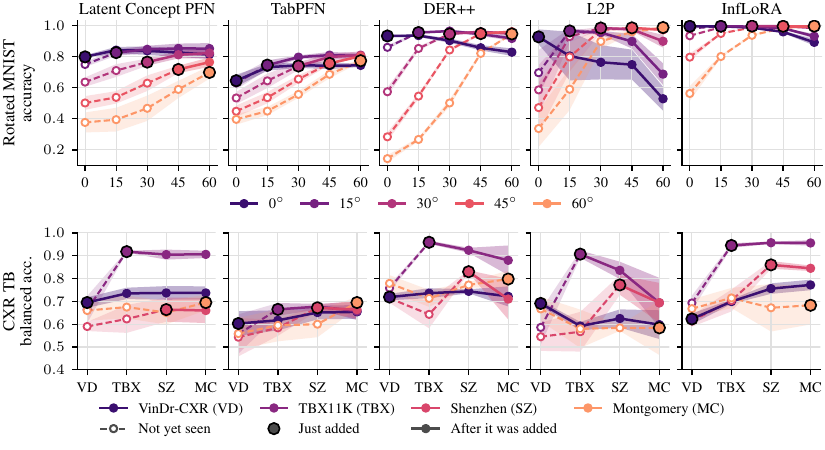}
\caption{Accuracy on each test domain as domains are added, on two domain-incremental streams: Rotated MNIST ($0^\circ \rightarrow 60^\circ$, top) and CXR TB (balanced accuracy, bottom), one panel per method. Dashed segments with open markers show forward transfer (domain not yet seen), ringed markers the step at which it is added, and solid segments retention afterwards. Mean $\pm 1$ sd over 5 seeds.}

\label{fig:domil-acc}
\label{fig:rotmnist-acc}
\label{fig:cxr_columns_by_method}
\end{figure}

\textbf{The Latent Concept PFN forgets least in every setting and stays competitive without a single parameter update.} On both domain-incremental benchmarks, forgetting is near-zero (FM $-0.007$ / BWT $+0.037$ on RotMNIST; FM $+0.005$ / BWT $+0.009$ on CXR), with per-task accuracy curves that are flat or slightly rising after introduction (Figs.~\ref{fig:rotmnist-acc}, \ref{fig:cxr_columns_by_method}, \ref{aa_metrics_table}). DER++ and InfLoRA reach higher AA ($0.918$/$0.777$ and $0.958$/$0.814$) by updating parameters on the full data at every step, and InfLoRA alone matches our low forgetting by constraining each step's update to a subspace orthogonal to previous steps' gradients. L2P forgets in every setting, likely because its shared prompt pool is overwritten by later steps. Our method makes no gradient updates, conditioning only on $k$ exemplars per class. Relative to TabPFN, the closest baseline, sharing our backbone, memory, and meta-training data, the AA gap is small on RotMNIST ($+0.004$) but larger on CXR ($+0.081$), suggesting that concept-anchored latents matter most when the shift involves prior shift (TB prevalence varies from 5.5\% to 50\% across hospitals) rather than pure covariate shift.

\textbf{The Latent Concept PFN outperforms baselines in the class-incremental setting.} On Split MNIST, the Latent Concept PFN achieves the highest AA of all methods, outperforming even parameter-updating baselines. L2P and InfLoRA are rehearsal-free, so each step's head is trained on the two new classes alone and inter-task boundaries must emerge implicitly. DER++ sees earlier classes via replay but must re-fit its shared head to a growing label set, risking drift on old boundaries. A context-based model sidesteps this entirely: new classes are incorporated by extending the memory, with no decision boundaries to relearn. Concept anchoring matters most here: the gap over TabPFN is largest on Split MNIST ($+0.156$), suggesting that a compositional latent space allows the model to reuse structure from earlier classes rather than building new representations from scratch, making a handful of exemplars sufficient to distinguish new classes from old ones. The errors that remain tend to be between classes that are close in concept space, which we examine below.

\begin{table}[t]
\caption{Continual learning metrics on the three datasets, computed from the mean accuracy matrices over 5 seeds. Note that FWT is not defined for class-incremental Split MNIST and CXR uses balanced accuracy. The top two rows are frozen and condition on the same exemplar memory while the bottom three update parameters on the full data of each step.}
\label{tab:cl_metrics}
\centering
\resizebox{\linewidth}{!}{%
\begin{tabular}{l cccc ccc cccc}
\toprule
 & \multicolumn{4}{c}{rotated mnist (domain-il)} & \multicolumn{3}{c}{split mnist (class-il)} & \multicolumn{4}{c}{cxr tb (domain-il)} \\
\cmidrule(lr){2-5} \cmidrule(lr){6-8} \cmidrule(lr){9-12}
method & aa $\uparrow$ & bwt $\uparrow$ & fm $\downarrow$ & fwt $\uparrow$
       & aa $\uparrow$ & bwt $\uparrow$ & fm $\downarrow$
       & aa $\uparrow$ & bwt $\uparrow$ & fm $\downarrow$ & fwt $\uparrow$ \\
\midrule
Latent Concept PFN (ours) & 0.789 & $+0.037$ & $-0.007$ & $+0.568$ & 0.794 & $-0.072$ & $+0.072$ & 0.749 & $+0.009$ & $+0.005$ & $+0.146$ \\
TabPFN                    & 0.785 & $+0.068$ & $-0.013$ & $+0.553$ & 0.638 & $-0.093$ & $+0.093$ & 0.668 & $+0.012$ & $+0.008$ & $+0.080$ \\
\midrule
DER++                     & 0.918 & $-0.032$ & $+0.039$ & $+0.742$ & 0.749 & $-0.273$ & $+0.273$ & 0.777 & $-0.065$ & $+0.074$ & $+0.224$ \\
L2P                       & 0.813 & $-0.193$ & $+0.193$ & $+0.787$ & 0.471 & $-0.418$ & $+0.418$ & 0.642 & $-0.128$ & $+0.128$ & $+0.079$ \\
InfLoRA                   & 0.958 & $-0.045$ & $+0.045$ & $+0.883$ & 0.714 & $-0.058$ & $+0.058$ & 0.814 & $+0.048$ & $+0.000$ & $+0.190$ \\
\bottomrule
\end{tabular}}
\label{aa_metrics_table}
\end{table}

\begin{figure}
\centering
\includegraphics[width=\linewidth]{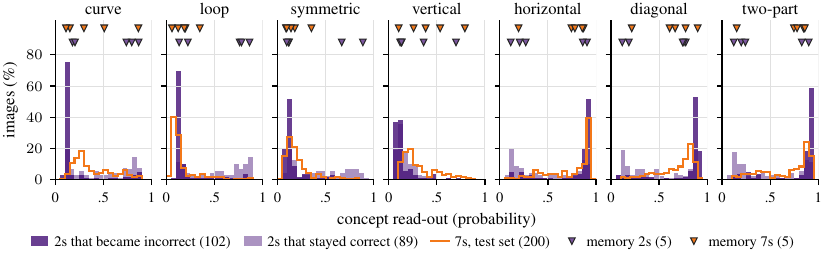}
\caption{Concept read-out on Split MNIST (seed 42) for 2s when 6 and 7 arrived. We show the 2s that became incorrect, and those that stayed correct alongside the concept-readout for the 7s in the test queries. Triangles mark the memory exemplars of 2 and 7.}
\label{fig:split-hist}
\end{figure}

\textbf{The concept read-out allows us to explain a drop in accuracy.} We demonstrate this on the largest drop observed across 5 seeds for Split MNIST: accuracy on digits 2 and 3 fell from $0.800$ to $0.560$ (seed 42) when 6 and 7 arrived, with 108 of 320 previously correct images flipping. Crucially, the concept read-out for these images changed by only $0.03$ per concept on average, and no more for the images that became incorrect than those that stayed correct ($0.025$ vs.\ $0.037$), confirming that the representation itself is stable when new classes arrive. The errors were concentrated: 92 of the 108 flips were 2s predicted as 7s. The lost 2s share a consistent concept profile, where at least 79\% read below $0.3$ on curve, loop and symmetric, and above $0.7$ on horizontal, diagonal and two-part: matching the code for the letter `z'. The stored 7 exemplars occupy the same region: three of the five read like a `z' on all six concepts. Of the 116 correctly classified 2s that also read like a `z', 72\% became incorrect after 7 arrived, against only 25\% of the other 2s. The concept read-out therefore identifies not just which images fail, but why: they are 2s whose concept code is indistinguishable from the newly arrived 7 exemplars (Fig.~\ref{fig:split-hist}).

\begin{figure}
  \centering
  \includegraphics[width=\linewidth]{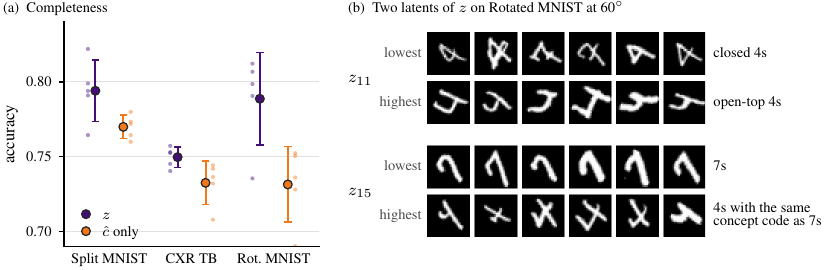}
  \caption{Concept completeness and what $z$ encodes beyond the concepts.
  (a) Accuracy at the end of the stream when stage 2 is given $z$ or only the concept read-outs $\hat{c}$, for both the memory exemplars and the test images (mean $\pm$ sd over 5 seeds)
  (b) Rotated MNIST test images at $60^\circ$ (seed 11) with the lowest and highest values of two latents of $z$. $z_{11}$ separates closed from open-top 4s, and $z_{15}$ separates 7s from 4s that share their concept code.}
  \label{fig:rot_completeness}
\end{figure}

\textbf{The concept set need not be complete as the latent $z$ carries information beyond the concepts.} We measure completeness by replacing $z$ with the concept read-outs $\hat{c}$ as input to stage 2 and adapting the completeness score of \citet{NEURIPS2020_ecb287ff} to measure the ratio of above-chance information in $\hat{c}$ relative to $z$. Scores are $0.97$, $0.93$, and $0.92$ on Split MNIST, CXR, and Rotated MNIST respectively: concepts carry most of the predictive signal, but a gap exists on all three datasets. We focus on RotMNIST, where the gap is largest. Of 25,000 test images, 2,120 are correct under $z$ but not under $\hat{c}$ alone, concentrated on digits 6 ($+0.11$), 3 ($+0.09$), and 5 ($+0.09$). The most common correction is $4\!\to\!7$ (108 images): open-topped 4s and nearly-vertical 7s share similar concept codes because the annotation assumes a diagonal stroke for 7 that many MNIST examples lack. To verify that $z$ encodes genuinely different information rather than merely shifting the boundary in concept space, we paired each such 4 with the test 7 at the same angle whose concept read-out was closest (mean difference $0.033$ per concept). $z$ classified both images in a pair correctly in 85\% of cases, whereas $\hat{c}$ alone called both a 7 in 88\% of pairs. Inspecting individual latent dimensions confirms this: one dimension separates open- from closed-topped 4s, and another separates those open-topped 4s from 7s, a variation in writing style that the predefined concept vocabulary does not capture (Fig.~\ref{fig:rot_completeness}).

\textbf{Where finding labels exist, we can confirm that the concept read-out is accurate.} On the 30,000 held-out VinDr films, the mean AUROC over twelve findings is $0.76$, reaching $0.89$ for cardiomegaly. For the other hospitals, where finding labels are unavailable, we instead examine cross-hospital consistency. Of the four findings with the largest spread in mean read-out across hospitals, pleural thickening and pulmonary fibrosis are consistently higher in TB-positive patients than TB-negative patients across all four hospitals. The gap is smallest for Shenzhen, which also has the lowest balanced accuracy at the end of the stream (Fig.~\ref{fig:cxr_concepts}, \ref{fig:cxr_columns_by_method}).

\begin{figure}[h]
\begin{center}
\includegraphics[width=\linewidth]{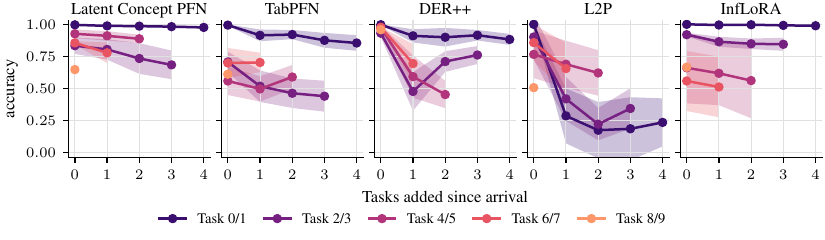}
\end{center}
\caption{Accuracy on each task of the class-incremental Split MNIST stream ((0,1) $\rightarrow$ (8,9)), one panel per method, aligned at the step the task is added: $x = k$ is accuracy after $k$ further tasks, so a falling line shows forgetting.}
\label{fig:splitmnist-acc}
\end{figure}

\section{Conclusion and Limitations}

We introduced the Latent Concept PFN, which reframes continual learning as continual inference: a model meta-trained with concept annotations to map inputs to a structured latent space $z$ and then to targets $y$, frozen thereafter, and adapted solely by conditioning on a growing exemplar memory. It achieves competitive AA against methods that train on the full data at every step, and outperforms all baselines on class-incremental Split MNIST. Concept anchoring is most beneficial under prior shift (CXR) and in the class-incremental setting. The concept read-out enables post-hoc error analysis, and $z$ encodes discriminative structure beyond the annotated concept vocabulary when needed. The primary limitation is that concept annotations are required at meta-training time, though label-free \citep{oikarinen2023labelfree} and post-hoc \citep{yuksekgonul2023posthoc} approaches suggest routes around this via vision-language models for example. Additionally, the concept read-out can help diagnose the source of errors, but it cannot yet intervene on the concepts to correct those errors, as in a conventional CBM.

\clearpage

\bibliography{references}

\bibliographystyle{plainnat}

\appendix

\section{Preliminaries}

\label{sec:preliminaries}
\subsection{Continual Learning}
Continual learning broadly concerns learning from a non-stationary stream of data \citep{vandeven2022three}. Three common scenarios are \citep{vandeven2022three}: (1) \emph{domain-incremental learning}, where the task remains fixed but the data domain changes; (2) \emph{class-incremental learning}, where new classes are introduced over time; and (3) \emph{task-incremental learning}, where a model learns multiple distinct tasks and the task identity is known at test time. Motivated by clinical prediction, we focus on the first two settings. For example, a model deployed across hospitals may encounter different imaging equipment, patient populations, or recording protocols while performing the same diagnostic task. Alternatively, a model initially trained to predict one disease may need to acquire the ability to predict additional diseases. In the latter setting, the disease to be predicted is not known a priori at test time, making task-incremental learning inappropriate.

Formally, a learner observes a stream of $T$ steps. At step $t$ it receives a dataset
$\mathcal{D}_t = \{(\vx^{(i)}, y^{(i)})\}_{i=1}^{n_t} \sim p_t(\vx, y)$ with label set $\mathcal{Y}_t$, and after step $t$ it is evaluated on all $p_{t'}$ with $t' \le t$, without access to $t'$.

\textbf{Domain-Incremental Learning}
The label space $\mathcal{Y}_t = \mathcal{Y} = \{1, \dots, K\}$ and the underlying task are shared across all steps, but the data distribution $p_t(\vx, y) = p_t(\vx)\, p_t(y \mid \vx)$ changes from one domain to the next. The change can be in the covariates, $p_t(\vx)$ (covariate shift), in the class prevalence, $p_t(y)$ (prior shift), or in the conditional $p_t(y \mid \vx)$ (concept shift). The goal is to maintain high accuracy on all domains seen so far after observing each new domain. Under concept shift this cannot be achieved by any fixed predictor $f(\vx)$ when the domain is not recoverable from $\vx$ alone, so predictions must also depend on stored examples.

\textbf{Class-Incremental Learning}
There are $K$ classes in total, but they arrive sequentially: at first we may only observe two or three classes, and new classes arrive over time. Each step introduces its own classes, so the label sets $\mathcal{Y}_t$ do not overlap and together make up $\{1, \dots, K\}$. After step $t$ the learner must distinguish between all classes seen so far, $\mathcal{Y}_1 \cup \dots \cup \mathcal{Y}_t$, without being told which step a test example came from. This is challenging because the decision boundaries change as new classes arrive, and classes from different steps are never observed together.

\textbf{Combined Domain and Class-incremental Learning.} It is often the case that we want to deal with both types of shift at once. For example, the COVID-19 pandemic
introduced a new diagnostic class for chest imaging while acquisition protocols, patient populations and disease
prevalence shifted at the same time \citep{roberts2021common, degrave2021ai}, and a deployed model is not told which type of
change has occurred.

\subsection{Prior-Data Fitted Networks}
During meta-training of PFNs \citep{muller2022transformers}, datasets
$\mathcal{D} \sim p(\mathcal{D})$ are sampled from a prior over datasets $p(\mathcal{D})$. For each sampled dataset,
one example $(\vx^{(q)}, y^{(q)})$ is held out as a query and the remaining examples form the context set
$\mathcal{D}_{\setminus q}$. The PFN $q_{\vtheta}$ is trained to approximate the posterior predictive distribution
$p(y^{(q)} \mid \vx^{(q)}, \mathcal{D}_{\setminus q})$ by minimizing the expected negative log-likelihood:
\begin{equation}
    \mathcal{L}(\vtheta) = \E_{\mathcal{D} \sim p(\mathcal{D})} \, \E_{(\vx^{(q)}, y^{(q)}) \sim \mathcal{D}}
    \left[ -\log q_{\vtheta}\big(y^{(q)} \mid \vx^{(q)}, \mathcal{D}_{\setminus q}\big) \right].
\end{equation}
Minimizing this loss is equivalent to minimizing the expected KL divergence from the posterior predictive
distribution to $q_{\vtheta}$, so a trained PFN approximates Bayesian inference under $p(\mathcal{D})$. At inference
time, labeled examples are provided as context tokens and the model predicts on new queries by conditioning on this
context through attention, with no parameter updates. Our method departs from the standard PFN formulation by keeping
a pretrained tabular PFN \cite{hollmann2023tabpfn, hollmann2025accurate} fixed as
the in-context classifier and meta-training only a map from inputs into a latent concept space that this classifier
reads, allowing continual adaptation through inference alone.

\section{Concepts}
\label{app:concepts}
 \begin{table}[h]
\caption{Concept annotation of the source dataset, EMNIST letters ($C = 7$). A 1 marks a concept that is present in the lowercase form of the letter. The 26 letters have 20 distinct codes; the letters that share a code are $\{a, o\}$, $\{c, u\}$, $\{h, n, r\}$, $\{i, l\}$ and $\{w, x\}$. During meta-training each bit is flipped independently with probability 0.1 whenever it is used.}
\label{tab:letter-concepts}
\centering
\resizebox{\linewidth}{!}{%
\begin{tabular}{l*{26}{c}}
\toprule
 & a & b & c & d & e & f & g & h & i & j & k & l & m & n & o & p & q & r & s & t & u & v & w & x & y & z \\
\midrule
curve      & 1 & 1 & 1 & 1 & 1 & 0 & 1 & 1 & 0 & 1 & 0 & 0 & 1 & 1 & 1 & 1 & 1 & 1 & 1 & 1 & 1 & 0 & 0 & 0 & 1 & 0 \\
vertical   & 0 & 1 & 0 & 1 & 0 & 0 & 0 & 1 & 1 & 0 & 1 & 1 & 1 & 1 & 0 & 1 & 1 & 1 & 0 & 0 & 0 & 0 & 0 & 0 & 0 & 0 \\
horizontal & 0 & 0 & 0 & 0 & 1 & 1 & 0 & 0 & 0 & 0 & 0 & 0 & 0 & 0 & 0 & 0 & 0 & 0 & 1 & 1 & 0 & 0 & 0 & 0 & 0 & 1 \\
diagonal   & 0 & 0 & 0 & 0 & 0 & 1 & 0 & 0 & 0 & 0 & 1 & 0 & 0 & 0 & 0 & 0 & 1 & 0 & 0 & 1 & 0 & 1 & 1 & 1 & 1 & 1 \\
loop       & 1 & 1 & 0 & 1 & 1 & 0 & 1 & 0 & 0 & 0 & 0 & 0 & 0 & 0 & 1 & 1 & 1 & 0 & 0 & 0 & 0 & 0 & 0 & 0 & 0 & 0 \\
two-part   & 0 & 1 & 0 & 0 & 0 & 0 & 0 & 0 & 0 & 0 & 1 & 0 & 1 & 0 & 0 & 0 & 0 & 0 & 1 & 0 & 0 & 0 & 1 & 1 & 0 & 1 \\
symmetric  & 1 & 1 & 1 & 1 & 1 & 0 & 0 & 0 & 1 & 0 & 1 & 1 & 1 & 0 & 1 & 0 & 0 & 0 & 0 & 0 & 1 & 1 & 1 & 1 & 0 & 0 \\
\bottomrule
\end{tabular}}
\end{table}

\begin{table}[h]
\caption{Concept annotation of the MNIST digits, with the same $C = 7$ concepts. It is used only to evaluate the concept read-out and is never given to the model. All ten digits have distinct codes.}
\label{tab:digit-concepts}
\centering
\begin{tabular}{l*{10}{c}}
\toprule
 & 0 & 1 & 2 & 3 & 4 & 5 & 6 & 7 & 8 & 9 \\
\midrule
curve      & 1 & 0 & 1 & 1 & 0 & 1 & 1 & 0 & 1 & 1 \\
vertical   & 0 & 1 & 0 & 0 & 1 & 1 & 0 & 0 & 0 & 1 \\
horizontal & 0 & 0 & 1 & 0 & 1 & 1 & 0 & 1 & 0 & 0 \\
diagonal   & 0 & 1 & 0 & 0 & 1 & 0 & 0 & 1 & 0 & 0 \\
loop       & 1 & 0 & 0 & 0 & 0 & 0 & 1 & 0 & 1 & 1 \\
two-part   & 0 & 0 & 0 & 1 & 0 & 0 & 0 & 0 & 1 & 0 \\
symmetric  & 1 & 1 & 0 & 1 & 0 & 0 & 0 & 0 & 1 & 0 \\
\bottomrule
\end{tabular}
\end{table}

\FloatBarrier

\section{Detailed Experimental Setup}
\label{app:exptsetup}

\textbf{Source dataset for the MNIST experiments} Both MNIST experiments are meta-trained on EMNIST letters \citep{cohen2017emnist}, $28 \times28$ grayscale images in the same format as MNIST. This consists of 26 classes, one per letter of the alphabet, where the letter could be upper or lowercase. There are no digits, so no class of either stream is seen during meta-training. Each letter has an annotation of $C = 7$ binary concepts that cover various stroke attributes such as curve or vertical stroke. The concepts were chosen such that they could apply to digits as well, and a latent space anchored to them during meta-training can carry
over to digits. The annotation is a property of the class rather than of the individual image similar to the protocol used in CBMs \citep{pmlr-v119-koh20a}. The concepts describe the lowercase form, so it is only approximate for the uppercase images of the same class, and each bit is additionally flipped independently with probability $0.1$ whenever it is used. It is also incomplete, there are 20 distinct codes for the 26 letters. The letters supply images, concept annotations and sometimes a label. The label of an episode is either the identity of the letter or drawn from rule prior in Section~\ref{sec:metatrain}. Note that concept annotations are used only during meta-training, and are not provided during the CL streams.

\textbf{Rotated MNIST} The domain-incremental stream is Rotated MNIST \citep{NIPS2017_f8752278}, built on the MNIST handwritten digits \citep{lecun1998gradientbased}. Our 5 steps are as follows: rotation by $0^\circ$, $15^\circ$, $30^\circ$,
$45^\circ$ and $60^\circ$, with bilinear interpolation applied to the image before normalisation. All ten digits appear in every domain and the label space is
shared, so the task is unchanged and only the observation process differs between
steps, which is covariate shift. At each step, we draw exemplars from the MNIST training split, rotate by that step's angle, and add it to the memory. For the queries, we draw from the the MNIST test split instead and rotate. After each step, we evaluate on every domain seen so far with the memory, which consists of a mix of exemplars from all domains seen so far, as the context. No domain id is given to the model and all ten classes are available from the first step onwards.

\textbf{Split MNIST} The class-incremental steam is Split MNIST \citep{pmlr-v70-zenke17a, vandeven2022three}, built on the same digits \citep{lecun1998gradientbased}. The five steps present
the disjoint pairs $(0,1)$, $(2,3)$, $(4,5)$, $(6,7)$ and $(8,9)$ in that order, so the label
sets do not overlap. The label set grows over our steps. After step $t$ the model must separate all $2t$ classes seen so far and is not told
which step a query came from. As in Rotated MNIST, the exemplars are drawn from the MNIST
training split and the queries from the MNIST test split, and after each step we evaluate on
every task seen so far. The memory grows by ten exemplars per step as we add just the two new digit exemplars for each step to the memory. Predictions are confined to classes seen so far.

\textbf{Source datasets for the chest x-ray task} The chest radiograph experiment is meta-trained on VinDr-CXR \citep{nguyen2022vindrcxr, PhysioNet-vindr-cxr-1.0.0, pollard_physionet_2026}, in which radiologists labelled every image with both local findings and global diagnoses. We use the local findings as our concepts, selecting $C=12$ findings with at least 100 positive images in the training split (e.g.\ cardiomegaly, pleural effusion) and taking the majority vote of the three readers of each training image. As we did for EMNIST, we flip each bit independently with probability 0.1. The diagnoses supply the labels. An episode picks one of five diagnoses (tuberculosis, pneumonia, lung tumour, other disease, no finding) and labels each image by whether it carries that diagnosis, or draws its label from the rule prior of Section~\ref{sec:metatrain}. We also add a slice of 1000 images from the official training split of TBX11L \citep{liu2020rethinking, liu2023revisiting}, the second hospital in the CL stream. These have a tubercolosis label but no findings, so their concept loss is masked, and they make up 10\% of episodes. These are disjoint from every TBX11K image the stream later uses. Unlike the MNIST experiments, the stream's label (tuberculosis) is one of the source diagnoses. All images are greyscale, resized to $224\times224$ and z-scored per image. The TabPFN baseline fits its PCA projection on the same 14500 source images, without labels, so it sees exactly the source data our model was meta-trained on.

\textbf{Chest x-ray CL stream} This is a domain-incremental stream as the label space does not grow. The stream consists of four hospitals: VinDR-CXR, TBX11K, Shenzhen \citep{jaeger2014two, candemir2014lung, jaeger2014automatic}, and Montgomery \citep{jaeger2014two, candemir2014lung, jaeger2014automatic}. We predict whether or not the scan received a label of tuberculosis. The hospitals differ in various ways such as the scanners available, patient populations and the prevalence of tuberculosis, which is 5.5\%, 11\%, 50\%, and 36\% respectively in the queries. We therefore use balanced accuracies for all metrics. The queries come from the following: The official VinDR-CXR test split (3000 images, labelled by consensus by 5 radiologists, the official validation split of TBX11K (1794 images), and a random half of Shenzhen (331) and of Montgomery (69). Note that the TBX11K validation split was used instead of test as the test labels are not public. Exemplars come from disjoint pools. At each step we store $k=16$ exemplars per class, so the memory holds 32 images per hospital and 128 after the last step. No hospital identity is given.
\begin{table}[t]
\caption{The three experiments. The two MNIST experiments use the same source dataset, EMNIST
letters \citep{cohen2017emnist}, and the same meta-trained model. The labels of the source
dataset are drawn from the prior rather than taken from
letter identity. Classes at each step counts the classes present both in the exemplars stored at
that step and in the queries it is evaluated on. $k$ is the number of exemplars stored per class
at each step, Eq.~\ref{eq:memory}; the memories end at different sizes because every domain of
Rotated MNIST contains all ten digits, while each step of Split MNIST introduces two new ones.}
\label{tab:datasets}
\begin{center}
\footnotesize
\setlength{\tabcolsep}{5pt}
\begin{tabular}{l|>{\centering\arraybackslash}p{0.20\linewidth}>{\centering\arraybackslash}p{0.20\linewidth}>{\centering\arraybackslash}p{0.20\linewidth}}
\toprule
 & {\bf Rotated MNIST} & {\bf Split MNIST} & {\bf Chest radiographs} \\
\midrule
Setting & domain-incremental & class-incremental & domain-incremental \\
\midrule
\multicolumn{4}{l}{\it Source dataset} \\
\midrule
Images                  & EMNIST letters & EMNIST letters & VinDR-CXR (train) + 1000 TBX11K (train) \\
Classes                 & 26 letters & 26 letters & 5 global diagnoses \\
Concepts $C$            & 7 stroke types & 7 stroke types & 12 radiological findings \\
\midrule
\multicolumn{4}{l}{\it Continual learning stream} \\
\midrule
Images                  & MNIST digits & MNIST digits & X-ray images from 4 hospitals \\
Steps ($T = 5$)         & $0^\circ$, $15^\circ$, $30^\circ$, $45^\circ$, $60^\circ$ & (0,1), (2,3), (4,5), (6,7), (8,9) & VinDr, TBX11K, Shenzhen, Montgomery \\
Classes at each step    & 10 & 2 & 2 \\
Exemplars per class $k$ & 5 & 5 & 16 \\
Memory after step $T$   & 250 & 50 & 128 \\
Queries per step        & 1000 & 400 & 3000, 1794, 331, 69 \\
\bottomrule
\end{tabular}
\end{center}
\end{table}

\section{Data}

\subsection{MNIST Benchmarks}

The meta-training prior for the MNIST benchmarks is built from EMNIST-letters, which contains 124,800 handwritten letter images across 26 classes. From this, 1,500 images per letter are randomly sampled (seed 0) and split 80/20 within each class, yielding a \textbf{training pool} of 1,200 images per letter (31,200 total) and a \textbf{held-out pool} of 300 images per letter (7,800 total). All meta-training gradient steps draw exclusively from the training pool; held-out images are reserved for evaluation episodes only.

\textbf{Inputs ($\mathbf{x}$).}
The 31,200 training-pool images are 28$\times$28 greyscale, normalised identically to MNIST. Shift episodes warp images on the fly at training time.

\textbf{Concepts ($\mathbf{c}$).}
Each letter class is assigned 7 binary concept features via a fixed, hand-coded lookup table, so every image of the same letter receives the same concept code. The seven concepts are: \textit{curve, vertical, horizontal, diagonal, loop, two\_part, symmetric}. Across the 26 letters, this yields 20 distinct codes; several letters share a code (a/o, c/u, h/n/r, i/l, w/x). The 10 digit classes each receive a unique code with no collisions. During training, each bit is independently flipped with probability 0.10 before being used as the binary cross-entropy (BCE) target.

\textbf{Labels ($\mathbf{y}$).}
Depending on the episode type, $\mathbf{y}$ is either the true letter identity or a synthetic label generated by the BNN rule (see Section~\ref{sec:method}).

\textbf{Data volume.}
Training runs for 8,000 steps with 32 episodes per step and 64 images per episode, giving 2,048 image draws per step and approximately 16.4 million draws in total, sampled with replacement across episodes.

\subsection{CXR Benchmark}

The prior for the CXR benchmark is built from VinDr-CXR~\citep{nguyen2022vindrcxr}, which contains 18,000 chest radiographs (15,000 official training, 3,000 official test), resized to 224$\times$224 pixels in greyscale. The official training split is itself divided using random seed 0 into $\mathcal{I}_\text{FIT}$ (90\%, 13,500 films) used for meta-training, and $\mathcal{I}_\text{HOLD}$ (10\%, 1,500 films) held out entirely. An additional 1,000 films are drawn from the official TBX11K~\citep{liu2020rethinking} training set (stratified by folder, seed 0; 93 positive for TB) and kept strictly disjoint from the TBX11K images used at test time.

\textbf{Inputs ($\mathbf{x}$).}
$13{,}500$ VinDr films from $\mathcal{I}_\text{FIT}$ plus 1,000 TBX11K films, giving 14,500 total.

\textbf{Concepts ($\mathbf{c}$).}
Twelve radiological findings that each appear in at least 100 positive cases within the VinDr training set are used as concept labels, annotated per image by majority vote of three radiologists. The findings are: \textit{Aortic enlargement, Calcification, Cardiomegaly, Consolidation, ILD, Infiltration, Lung Opacity, Nodule/Mass, Pleural effusion, Pleural thickening, Pulmonary fibrosis,} and \textit{Other lesion}. As with MNIST, each bit is randomly flipped with probability 0.10 for the BCE target. Concept labels are available only for the 13,500 VinDr films; the concept loss is masked on the TBX11K slice.

\textbf{Labels ($\mathbf{y}$).}
Each episode is framed as a binary classification: one of five diagnoses versus all others. Positive-class counts within $\mathcal{I}_\text{FIT}$ are: TB (436), Pneumonia (432), Lung tumor (118), Other disease (3,596), and No finding (9,542). For the TBX11K slice, only the TB label is used.

\textbf{Data volume.}
Training runs for 2,000 steps with 32 episodes per step and 64 images per episode, giving approximately 4.1 million total image draws.

\subsection{Continual Learning Streams}

The continual learning (CL) streams are used solely for evaluation. The model's weights remain \textbf{frozen} throughout; only a small set of exemplar images is stored in memory after each task or domain arrives.

\subsubsection{Split MNIST (Class-Incremental, 5 Tasks)}

Digit pairs arrive sequentially: $\{0,1\}$, $\{2,3\}$, $\{4,5\}$, $\{6,7\}$, $\{8,9\}$. After each task, 5 exemplars per class are stored from the MNIST training set (10 per task, building to 50 after all five tasks). Query sets contain 200 images per class from the MNIST test set (400 per task, 2,000 in total). The model receives no task identity at inference time and restricts its predictions to classes seen so far via argmax.

\subsubsection{Rotated MNIST (Domain-Incremental, 5 Domains)}

All 10 digit classes appear in every domain, but each domain applies a different rotation to the images: $0^\circ$, $15^\circ$, $30^\circ$, $45^\circ$, $60^\circ$ (bilinear interpolation). Memory stores 5 exemplars per class per domain (50 per domain, building to 250 after all five domains). Query sets contain 100 images per class per domain from the MNIST test set (1,000 per domain, 5,000 in total). The headline evaluation uses the entire accumulated memory as the in-context set. A secondary \textit{matched} protocol averages results over 8 random class-balanced subsets of at most 50 exemplars each.

\subsubsection{CXR TB (Domain-Incremental, 4 Hospitals)}

Hospitals arrive in sequence: VinDr $\to$ TBX11K $\to$ Shenzhen $\to$ Montgomery. The task is always binary: TB versus non-TB. After each hospital, 32 films are stored as exemplars (16 TB, 16 non-TB), building to 128 after all four hospitals; the entire accumulated set serves as the in-context memory. The primary evaluation metric is balanced accuracy; AUROC of the predicted TB probability is reported alongside. The model receives no hospital identity at inference time.

The CXR TB stream passes through four hospitals in sequence.
For VinDr, the memory pool is drawn from $\mathcal{I}_\text{HOLD}$
(1,500 films, 46 TB-positive), the held-out split that was never
seen during meta-trainingand the query set is the official VinDr
test set (3,000 films, 164 TB-positive, 5.5\% prevalence).
For TBX11K, memory is drawn from the official training set minus the
meta-training slice (5,480 films, 507 TB-positive), with queries taken
from the official validation set (1,794 films, 200 TB-positive, 11.1\%
prevalence).
For Shenzhen and Montgomery, each dataset is split in half by random
seed 0: the first half serves as the memory pool and the second half
as the query set.
Shenzhen contributes 331 films per split (171 and 165 TB-positive
respectively, 49.8\% prevalence in queries), and Montgomery contributes
69 films per split (33 and 25 TB-positive respectively, 36.2\%
prevalence in queries).

\section{Method Implementation}
\label{app:method}

The model is a three-stage pipeline: (1) a convolutional \textbf{image embedder} that maps each raw image to a 128-dimensional vector; (2) a \textbf{Stage~1 in-context transformer} that processes a full episode and produces a latent variable $\mathbf{z}$ per image; and (3) a fully frozen pretrained tabular classifier, \textbf{TabPFN}, that takes $\mathbf{z}$ as input and produces class predictions. A lightweight \textbf{concept head} additionally reads from $\mathbf{z}$ to predict concept labels.

\subsection{Components Shared Across All Datasets}

\textbf{Stage 1: In-context transformer.}
Stage~1 consists of 3 post-LayerNorm transformer blocks with hidden size 256, 4 attention heads, a 4$\times$ feed-forward expansion, GELU activations, and zero-initialised output projections. The attention mask follows the PFN convention: context rows attend to each other; query rows attend to context rows; query rows never attend to each other. The embedder's 128-dimensional output is projected to 256 dimensions by a linear layer before entering Stage~1. The combined front-end has 2.41M parameters.

The latent variable $\mathbf{z}$ is 16-dimensional and is drawn from a diagonal Gaussian:
\begin{equation}
    q(\mathbf{z} \mid \mathbf{x}, \mathcal{X}_\text{ctx}) = \mathcal{N}\!\left(\boldsymbol{\mu},\, \mathrm{diag}\!\left(\exp(\boldsymbol{\ell})\right)\right),
\end{equation}
where $\boldsymbol{\ell}$ is the predicted log-variance, clamped elementwise to $[-4, 4]$. No KL regularisation term is included in the loss.

\textbf{Concept head.}
A single linear layer maps from $\mathbf{z}$ (dimension 16) to $C$ concept outputs, followed by a sigmoid. BCE loss is computed on both context and query rows.

\textbf{Training objective.}
The loss combines a predictive term (negative log-likelihood under TabPFN) with a concept prediction term:
\begin{equation}
    \mathcal{L} = -\log\!\left(\frac{1}{S}\sum_{s=1}^{S} g_{\text{TabPFN}}\!\left(y_q \mid \mathbf{z}^{(s)}, \mathcal{X}_\text{ctx}\right)\right) + \lambda_c \cdot \mathrm{BCE}(\hat{\mathbf{c}}, \mathbf{c}),
\end{equation}
where $S=4$ latent samples are drawn jointly during training and $\lambda_c = 1$ in all headline results.

\textbf{Episode structure.}
Each episode contains 64 rows. The number of context rows $n_\text{ctx}$ is drawn uniformly from $\{8, \ldots, 56\}$ and is shared across all episodes in a batch; the remaining rows serve as queries. In 30\% of episodes, class counts are balanced; otherwise, exactly one image per class is placed in context and the rest are assigned at random. Each class is guaranteed at least one context row. A random class-to-slot mapping is drawn independently for each episode.

\textbf{Stage 2: TabPFN.}
We use the raw \texttt{PerFeatureTransformer} from \texttt{tabpfn==2.0.9}, a pretrained tabular classifier with 7.24M parameters, 12 transformer layers, and embedding size 192. All weights are completely frozen throughout training and evaluation; only a single forward pass is performed (no ensembling). Class labels are passed in as ranks of the classes present, sorted by class ID. Per-layer activation recomputation is disabled during training.

\textbf{Prior mixture.}
Each episode is generated as one of three types. With probability 0.40 the episode is a \textsc{Shift} episode; otherwise, with probability 0.70, it is a \textsc{Rule} episode; otherwise it is an \textsc{Identity} episode. This gives approximate marginal probabilities of 0.40, 0.42, and 0.18 respectively.

\begin{itemize}
    \item \textbf{Rule episodes.} The synthetic label is generated as
    \begin{equation}
        y = \mathrm{MulticlassRank}\!\left(\mathrm{BNN}\!\left([2\mathbf{c} - \mathbf{1};\, s(\mathbf{x} - \bar{\mathbf{x}})]\right)\right),
    \end{equation}
    where the BNN architecture and hyperprior follow the non-causal branch of TabPFN. The mixing coefficient $\rho$ is set to 0 with probability 0.25 (concept-only rule, no pixels), otherwise drawn from $\log$-Uniform$[0.1, 10]$, controlling the fraction of pixel variance mixed into the input. Up to 10 resamples are performed until every query label appears at least once in the context. The \textit{clean} concept values feed the BNN rule; the \textit{noise-corrupted} concept values are used as the BCE target.

    \item \textbf{Shift episodes.} The number of warp domains per episode is $K \sim \mathrm{Uniform}\{1, \ldots, 5\}$. Each domain applies a smooth spatial warp defined by a BNN coordinate field on a $7{\times}7$ grid of control points, bilinearly interpolated to image resolution, with the mean displacement subtracted to prevent global translation. Warp strength $\alpha$ is 0 with probability 0.25; otherwise $\alpha \sim \log\text{-Uniform}[0.05, 0.5] \times \bar{r}$, where $\bar{r}$ is the mean image radius. With probability 0.30, one domain is withheld from the context, requiring the model to generalise to an unseen warp. Labels and concepts are always those of the original, unwarped image.
\end{itemize}

\textbf{Optimiser and schedule.}
We use AdamW with learning rate $3 \times 10^{-4}$ and weight decay $10^{-4}$. The learning rate warms up linearly over the first 10\% of steps, then follows a cosine decay to zero. Gradient norms are clipped to 1.0. Each update step accumulates gradients over 32 episodes via micro-batches.

\textbf{Evaluation protocol.}
All model weights are frozen. At test time, $S_\text{eval} = 128$ latent samples are drawn and their predicted class probabilities are averaged. The entire stored memory (all accumulated exemplars) is used as the in-context set.

\textbf{Experimental seeds and metrics.}
All experiments are repeated with 5 random seeds (42, 123, 7, 11, 19); results are reported as mean $\pm$ sample standard deviation. Let $a_{ij}$ denote accuracy on task $j$ after training on tasks $1, \ldots, i$. We report:
\begin{itemize}
    \item \textbf{AA} (Average Accuracy): mean of the final row $a_{T,j}$, measuring overall performance at the end of the stream.
    \item \textbf{BWT} (Backward Transfer): $\frac{1}{T-1}\sum_{j<T}(a_{T,j} - a_{j,j})$, measuring how much performance on past tasks has dropped since they were first encountered.
    \item \textbf{FM} (Forgetting Measure,~\citealt{chaudhry2018riemannian}): similar to BWT but relative to the maximum accuracy ever achieved on each task (the final task is excluded from this maximum).
    \item \textbf{FWT} (Forward Transfer): average accuracy on a new domain before any of its exemplars are observed, relative to chance; reported for domain-IL benchmarks only.
\end{itemize}

\subsection{MNIST-Specific Settings}

The same trained model is used for both Split MNIST and Rotated MNIST. The image embedder is a three-layer CNN: three convolutional layers with 32, 64, and 64 channels (3$\times$3 kernels, ReLU activations), followed by two rounds of $2\times$ max-pooling and global average pooling. A final linear layer maps the 64-dimensional pooled representation to 128 dimensions. The embedder has approximately 64k parameters.

Stage~1 uses no dropout. Training runs for 8,000 steps using gradient accumulation over 2 micro-batches of 16 episodes each. Episodes may involve between 2 and 10 classes (the maximum supported by TabPFN). For Rule episodes, the BNN input consists of the 7 concept bits concatenated with all 784 pixel values.

\subsection{CXR-Specific Settings}

The image embedder is a ResNet-18 trained from scratch with a single input channel. Each image is normalised per-image using z-score normalisation (no dataset-level statistics). The ResNet output is projected to 128 dimensions and has approximately 11.2M parameters. BFloat16 mixed-precision (\texttt{bf16}) is used inside the ResNet to reduce memory usage. No data augmentation is applied.

Stage~1 uses dropout of 0.1. Training runs for 2,000 steps (the final checkpoint is used), with gradient accumulation over 8 micro-batches of 4 episodes each. Training was performed on a single A100 (40\,GB), at approximately 2 seconds per step. All episodes are binary (one diagnosis versus all others). Ten percent of episodes are drawn from the TBX11K slice, using only the TB label with the concept loss masked.

For Rule episodes, the BNN input is the 12 radiological findings concatenated with the image average-pooled to $28\times28$ (784 values) and z-scored. Rule episodes are always binary; TBX11K-slice episodes are never used as Rule episodes. The spatial warp field uses the same settings as MNIST, applied at $224\times224$ resolution.

At inference time, the predicted class is the argmax over \{non-TB, TB\}. For AUROC computation, the TB probability is renormalised over these two classes only.

\section{Baselines Overview}
\label{app:baselines_overview}
We compare against one replay method, two pretrained-backbone continual learning methods and a standard TabPFN. Aside from the TabPFN and our own method, the other baselines update parameters on the stream.

i) \emph{DER++} \citep{NEURIPS2020_b704ea2c} interleaves replayed exemplars with the current step's data and adds a distillation term. Its replay buffer is set to the size of our exemplar memory (50 for Split MNIST, 250 for Rotated MNIST), so
the two methods store the same number of examples. However, DER++ performs gradient updates on that buffer while we only use it as context.
ii) \emph{L2P} \citep{Wang_2022_CVPR} maintains a frozen pre-trained backbone. It learns from a pool of prompts, selecting a subset per input via a key-query match, so adaptations are confined to the prompt parameters.
iii) \emph{InfLoRA} \citep{liang2024inflora} adds a low-rank branch per task, constrained to a subspace orthogonal to the gradient space of previous tasks, to a frozen pre-trained backbone.
iv) \emph{TabPFN} We pass each image through a frozen ImageNet-pretrained ResNet-18 (cite), get z-scores for its 512-d pooled features and finally project them to 16 dimensions (same size as our latent $z$) using PCA. PCA and z-score are fitted once, without labels, on the same images that we use for meta training. This is then used as input into the same frozen TabPFN that we use for stage 2, using the same memories as our method for context.

Both L2P and InfLoRA use a ViT-B/16 backbone supervised-pretrained on ImageNet-21k and are
trained for five epochs per task on the full data of each step. Our method and TabPFN are
meta-trained on EMNIST letters and are then frozen, conditioning on only the
$k = 5$ exemplars per class held in memory.

\section{Baseline Implementation Details}
\label{app:baselines}

All baselines use the same five random seeds (42, 123, 7, 11, 19), the same query sets, and the same evaluation metrics as our method. No task or domain identity is provided at training or test time.

\subsection{TabPFN on Frozen ImageNet Features}

This baseline replaces our learned embedder with a frozen ResNet-18 pretrained on ImageNet (\texttt{IMAGENET1K\_V1} weights from \texttt{torchvision}), operating in evaluation mode throughout. The 512-dimensional global-average-pooled features are extracted without any fine-tuning.

\textbf{Input preprocessing.}
Images are resized to $224\times224$ pixels using bilinear interpolation without centre-cropping, and normalised using the ImageNet channel mean and standard deviation. For MNIST, the MNIST-specific normalisation is first undone before resizing from $28\times28$ to $224\times224$. For CXR, the 224-pixel greyscale films are divided by 255. In both cases, greyscale images are replicated across three channels.

\textbf{Dimensionality reduction.}
The 512-dimensional features are z-scored and then reduced to 16 dimensions via PCA, fitted once without labels on the same image pool used for our embedder's meta-training:
\begin{itemize}
    \item \textbf{MNIST}: the EMNIST-letters training pool (31,200 images); PCA-16 explains 52.0\% of the variance.
    \item \textbf{CXR}: the VinDr $\mathcal{I}_\text{FIT}$ set plus the TBX11K slice (14,500 films); PCA-16 explains 49.5\% of the variance.
\end{itemize}

\textbf{Classifier.}
We use \texttt{TabPFNClassifier} (\texttt{tabpfn==2.0.9}) with \texttt{random\_state} set to the current seed and \texttt{n\_estimators} = 4 for MNIST and 8 for CXR. The context passed to TabPFN is exactly the stored memory used by our method: 50 exemplars at the end of Split MNIST, the full union of 250 exemplars for Rotated MNIST, and the full union of 128 films for CXR TB. The classifier requires no gradient-based training; it is re-fitted on the current memory at every evaluation step.

\subsection{DER++}

DER++ augments standard experience replay with a distillation term that matches current logit outputs on buffer images to the logits stored at insertion time. We use a standard reservoir buffer and apply the method to both MNIST and CXR.

\textbf{Loss.}
Each training step minimises:
\begin{equation}
    \mathcal{L} = \mathcal{L}_\text{CE}(\text{stream batch}) + 0.5 \cdot \mathrm{MSE}\!\left(f(\mathbf{x}_\text{buf}),\, \hat{\mathbf{y}}_\text{buf}\right) + 0.5 \cdot \mathcal{L}_\text{CE}(\text{buffer batch}),
\end{equation}
where $\hat{\mathbf{y}}_\text{buf}$ are the logits stored at insertion time. Two independent buffer draws are used per step.

\textbf{Optimiser.}
AdamW with learning rate $3\times10^{-4}$, weight decay $10^{-4}$, gradient norm clipping to 1.0, and batch size 128. Training runs for 15 epochs per task/domain/hospital, with a single optimiser maintained across the entire stream. The buffer capacity is set to match the total memory size used by our method at the end of the stream.

\textbf{MNIST (Split and Rotated).}
The backbone is our MNIST CNN embedder architecture (3-layer CNN, identical structure, trained from scratch) with an additional \texttt{Linear(128, 10)} classification head. No data augmentation is applied. Training data consists of 1,000 MNIST training images per class per task, giving approximately 2,000 images per task for Split MNIST (240 steps per task) and 10,000 rotated images per domain for Rotated MNIST (1,185 steps per domain). The buffer holds 50 exemplars for Split MNIST (10 classes $\times$ 5 exemplars) and 250 for Rotated MNIST (10 classes $\times$ 5 exemplars $\times$ 5 domains). Each step processes 128 stream images plus two independent draws of $\min(128,\,|\text{buffer}|)$ buffer images. At inference, the argmax is restricted to classes seen so far, matching our method.

\textbf{CXR.}
The backbone is our CXR ResNet-18 (single input channel, trained from scratch, 128-dimensional output) with an additional \texttt{Linear(128, 2)} head. Data augmentation is applied on every training pass: random rotation of $\pm10^\circ$, scale $\pm10\%$, and shift $\pm8\%$; BatchNorm layers are kept in training mode throughout. Training data per hospital is:
\begin{itemize}
    \item VinDr: 15,000 films ($\mathcal{I}_\text{FIT} + \mathcal{I}_\text{HOLD}$), $\approx$1,770 steps.
    \item TBX11K: 6,480 films (5,480 from the training set plus the 1,000-film meta-training slice), $\approx$765 steps.
    \item Shenzhen: 331 films, $\approx$45 steps.
    \item Montgomery: 69 films, $\approx$15 steps.
\end{itemize}
Class-balanced sampling with replacement is used throughout (VinDr is approximately 3\% TB positive). The buffer holds 128 films total (32 per hospital $\times$ 4 hospitals); reservoir sampling determines which films are retained as the stream progresses.

\subsection{L2P and InfLoRA (CXR Only)}

Both Learning to Prompt and InfLoRA~\citep{liang2024inflora} use a ViT-B/16 backbone pretrained on ImageNet-21k (\texttt{vit\_base\_patch16\_224.augreg\_in21k} from \texttt{timm}), which remains completely frozen throughout. Greyscale CXR images are resized to $224\times224$, replicated across three channels, and scaled to $[0,1]$; ImageNet normalisation is not applied. Training data per hospital matches the DER++ CXR setup above. No rehearsal buffer is used.

\textbf{L2P.}
A prompt pool of $M=10$ prompts of length $L_p=5$ tokens is maintained. At each forward pass, the top $N=5$ prompts are selected by key-query matching, with key-matching loss weight $\lambda=0.5$. The trainable parameters are the prompts, the matching keys, and a \texttt{Linear(768, 2)} classification head. Training uses Adam ($\beta_1=0.9$, $\beta_2=0.999$) at learning rate $3\times10^{-3}$ ($1\times10^{-3}$ for split mnist), batch size 128, fp16 AMP, and cross-entropy loss, for 5 epochs per hospital with a single optimiser maintained across the stream.

\textbf{InfLoRA.}
An InfLoRA branch of rank $r=10$ is applied to the \texttt{qkv} projection of every transformer block (12 in total). After each hospital, the matrix $B_t$ is set to the top-$r$ left singular vectors of the layer inputs after projecting out the accumulated DualGPM subspace; only the matrix $A_t$ is trained. At the end of each hospital, $A_t B_t$ is merged into the frozen weights and the DualGPM subspace is updated. The DualGPM threshold increases linearly: $\varepsilon_t = 0.95 + 0.05 \cdot t/T$ for hospital $t \in \{1,\ldots,T\}$. The subspace is computed from 32 randomly sampled training images per hospital (seeded). The trainable parameters are $A_t$ and a \texttt{Linear(768, 2)} head; a fresh Adam optimiser ($\beta_1=0.9$, $\beta_2=0.999$) is used per hospital at learning rate $5\times10^{-4}$, batch size 128, fp16 AMP, and cross-entropy loss, for 5 epochs per hospital.

\section{Evaluation}
\label{app:evaluation_metrics}
\textbf{Interpreting the latent space}
The concept network $h_\phi$ (Section~4.2) gives an independent probability for each of the concepts. Since $h_\phi$ is trained on posterior draws of $z$ (Eq.~(8)), we marginalise the latent in the same way as the label predictive (Eq.~(5)). For a query $x^{(q)}$ with the memory $\mathcal{M}_t$ as context,
\[
\hat{c}^{(q)} \;=\; \mathbb{E}_{z \sim q_\theta(\cdot \mid x^{(q)}, X_{\mathrm{ctx}})}\big[h_\phi(z)\big] \;\approx\; \frac{1}{S}\sum_{s=1}^{S} h_\phi\big(z^{(q,s)}\big),
\]
reusing the $S$ draws that give the class prediction, so that $\hat{c}^{(q)}_k$ is the predicted probability that concept $k$ is present. To evaluate the read-out, we annotate the ten digits with the same seven concepts (Table~\ref{tab:digit-concepts}). These annotations are used only for scoring and never reach the model. We score each concept by its AUROC against the annotation. When describing where an input lies in latent space, for example how far its latent moves when a step arrives, or which memory exemplar or held-out letter image (never used in meta-training, with its own letter episode as context) is nearest to it, we use the posterior mean $\mu_\theta(x^{(q)}, X_{\mathrm{ctx}})$. For CXR, we only have concept annotations for the VinDr-CXR set.

\textbf{Metrics} We use standard continual learning metrics computed over the mean accuracy $aij$ matrices. i) Average accuracy (AA) \citep{NIPS2017_f8752278}, computed as the mean of the final row in the $aij$ matrices. $\mathrm{AA} \in [0, 1]$, higher is better. ii) Backward transfer (BWT) \citep{NIPS2017_f8752278}, where $\mathrm{BWT} \in [-1, 1]$, higher is better. iii) Forgetting Measure (FM) \citep{chaudhry2018riemannian} with the final step excluded from the maximum, where $\mathrm{FM} \in [-1, 1]$, lower is better. iv) Forward Transfer (FWT) \citep{NIPS2017_f8752278}, $\mathrm{FWT} \in [-\tfrac{1}{C},\, 1 - \tfrac{1}{C}]$, higher is better. We do not measure this for Split MNIST as in the class incremental case, the model has not been trained to output $j$'s classes before task $j$ arrives. FWT usually subtracts the accuracy of a randomly initialised model. As our method and most of our baselines build on pretrained components, we instead subtract chance ($1/classes$ or $0.5$ for balanced accuracy in cxr).

\section{Additional Figures}
\begin{figure}[t]
    \centering
    \includegraphics[width=\linewidth]{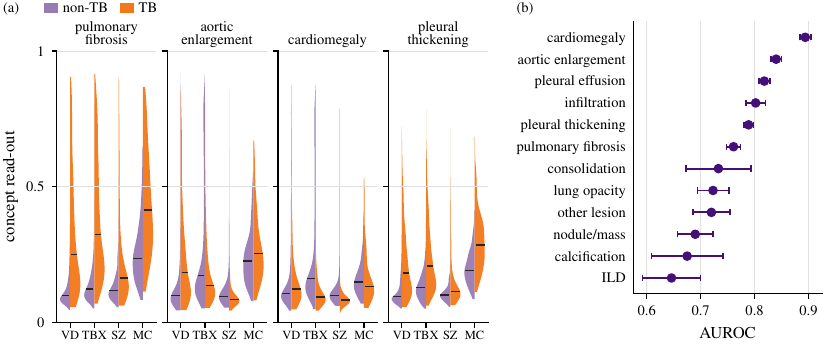}
    \caption{Concept read-out on the chest x-ray stream. (a) distribution of the read-out $\hat c$ over the query films of each hospital, split by tuberculosis label, for the four findings whose mean read-out differs most across hospitals. (b) read-out AUROC per finding on the vindr-cxr test split mean $\pm$ sd over five seeds.}
    \label{fig:cxr_concepts}
\end{figure}

\begin{figure}[t]
\centering
\includegraphics[width=\linewidth]{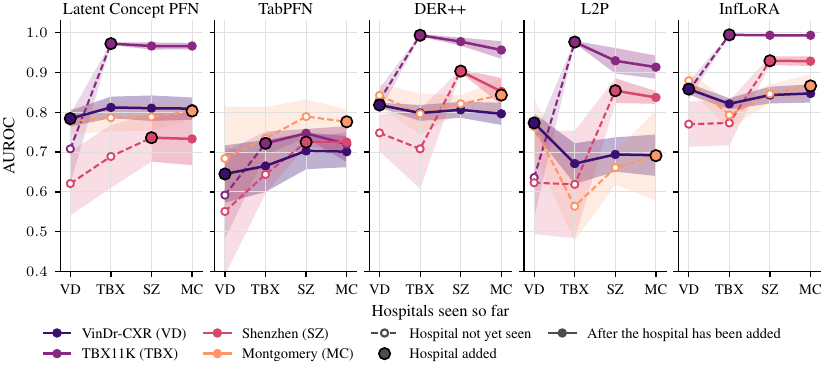}
\caption{AUROC on each test hospital along the cxr tb stream, one panel per method (mean $\pm$ sd over 5 seeds). balanced-accuracy version in figure~\ref{fig:cxr_columns_by_method}.}
\label{fig:cxr_auroc_columns_by_method}
\end{figure}

\begin{figure}[t]
\centering
\includegraphics[width=\linewidth]{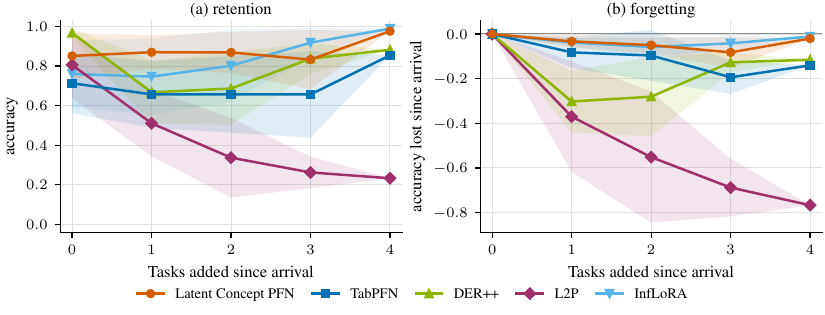}
\caption{Retention on the class-incremental Split MNIST stream, as a function of the number of tasks added since a task arrived. (a) Accuracy on the task; (b) accuracy lost relative to the step at which the task arrived, so each task's own difficulty cancels out. Over 5 seeds; shaded bands are $\pm 1$ sd across those tasks (a single task at $k = 4$).}
\label{fig:split-retention-forgetting}
\end{figure}
\FloatBarrier
\end{document}